\documentclass[lettersize,journal]{IEEEtran}
\usepackage{amsmath,amsfonts}
\usepackage{algorithmic}
\usepackage{algorithm}
\usepackage{array}
\usepackage[caption=false,font=normalsize,labelfont=sf,textfont=sf]{subfig}
\usepackage{textcomp}
\usepackage{stfloats}
\usepackage{url}
\usepackage{verbatim}
\usepackage{graphicx}
\usepackage{cite}
\usepackage{tabularx,booktabs}
\usepackage{multirow}
\usepackage{adjustbox}
\usepackage[table]{xcolor}
\usepackage{tabularray}
\usepackage[shortlabels]{enumitem}

\begin{document}

\title{Unsupervised Sim-to-Real Adaptation of Soft Robot Proprioception using a Dual Cross-modal Autoencoder}
\author{Chaeree Park, Hyunkyu Park, and Jung Kim
\thanks{Chaeree Park, Hyunkyu Park, Jung Kim are with the Department of Mechanical Engineering, Korea Advanced Institute of Science and Technology, Republic of Korea (e-mail: cofl0530, hkpark93, jungkim@kaist.ac.kr).}}

\maketitle

\begin{abstract}

Soft robotics is a modern robotic paradigm for performing dexterous interactions with the surroundings via morphological flexibility. The desire for autonomous operation requires soft robots to be capable of proprioception and makes it necessary to devise a calibration process. These requirements can be greatly benefited by adopting numerical simulation for computational efficiency. However, the gap between the simulated and real domains limits the accurate, generalized application of the approach. Herein, we propose an unsupervised domain adaptation framework as a data-efficient, generalized alignment of these heterogeneous sensor domains. 
A dual cross-modal autoencoder was designed to match the sensor domains at a feature level without any extensive labeling process, facilitating the computationally efficient transferability to various tasks. As a proof-of-concept, the methodology was adopted to the famous soft robot design, a multigait soft robot, and two fundamental perception tasks for autonomous robot operation, involving high-fidelity shape estimation and collision detection. The resulting perception demonstrates the digital-twinned calibration process in both the simulated and real domains. The proposed design outperforms the existing prevalent benchmarks for both perception tasks. This unsupervised framework envisions a new approach to imparting embodied intelligence to soft robotic systems via blending simulation.

\end{abstract}


\section{Introduction}
\IEEEPARstart{S}{oft} robots, composed of soft and stretchable materials, have long inspired future engineering applications towards safe, adaptive, and resilient interactions with unstructured environments and living organisms \cite{justus2019biosensing, wang2018toward, shih2020electronic, rus2015design, rothemund2021shaping}. Unlike traditional rigid robots, the inherent mechanical compliance of soft robots offers conformability and robustness to physical contact, which in turn comes at the cost of vulnerability. Therefore, the successful operation of autonomous soft robots demands delicate proprioception, which refers to the capability to intrinsically sense its own body kinematics, mainly using soft, stretchable sensors. These soft sensors adopt extensive stimuli-responsive materials, such as liquid metals \cite{park2010hyperelastic}, conductive nanocomposites \cite{yamada2011stretchable}, and permanent magnets \cite{ozel2015precise} to perform various functionalities on demand. However, the accurate modeling of the kinematics analytically and numerically using these soft sensors is challenging owing to inconsistent manufacturing, viscoelastic hysteresis, and high nonlinearities in their dynamics \cite{thuruthel2019soft}.

\begin{figure}[t]
\begin{center}
\includegraphics[width=\columnwidth] {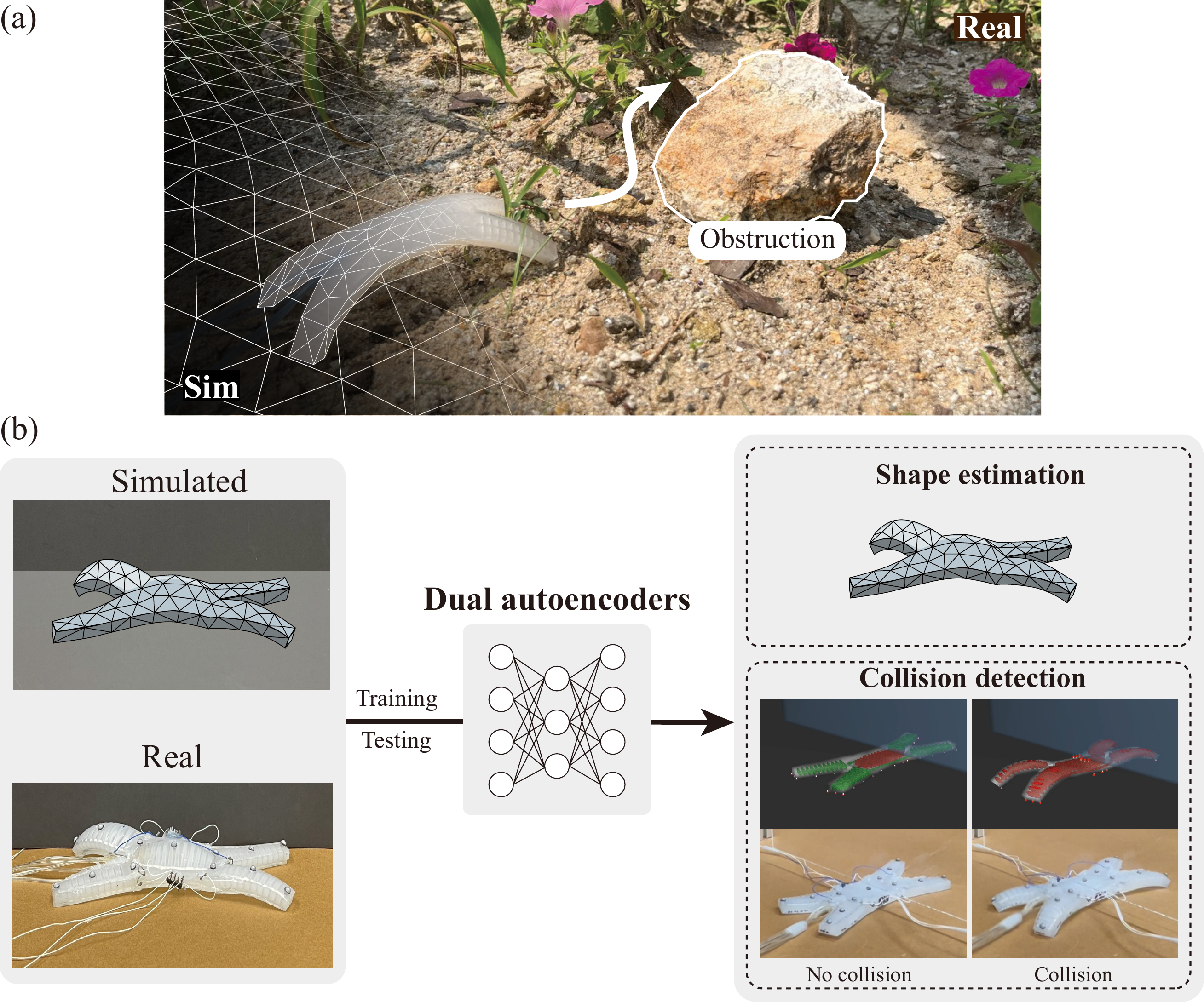}

\caption{\label{Main} Concept of our sim-to-real adaptation framework for proprioceptive soft robots. (a) Conceptual illustration of the soft robot with sim-to-real learning-based multimodal control. (b) The proposed network of domain adaptation with shape estimation and collision detection in the simulated and real domains. Simulation provides a high-fidelity mesh representation for shape estimation, while the real world relies on marker positions. In collision detection, simulation allows for generating labeled datasets more easily compared to the real world.}
\end{center}
\end{figure}

Machine learning methods have shown great success in overcoming these limitations \cite{kim2021review, chin2020machine}. Such data-driven approaches circumvent the explicit formulation of complicated, redundant soft robot dynamics. End-to-end mapping by embedded soft proprioceptive sensors is extensively leveraged for robot shape estimation \cite{truby2022fluidic, truby2020distributed, loo2022robust, van2018soft}, tactile sensing \cite{han2018use}, object identification \cite{shih2017custom}, and motion control \cite{thuruthel2018model}. However, current achievements suffer from inefficiencies in data acquisition as soft robot production varies largely depending on the manufacturing technique.
In addition, the experimental process for the explicit representation of the robot shape mainly relies on optical camera measurements. Because of the use of optical markers, which require a certain gap among them, the explicit representation is confined to low-quality data, and visual occlusion occurs during large deformations. Considering these problems, sim-to-real approaches, which have been widely used in the field of generic robotics, are regarded as alternatives for optical measurements \cite{graule2022somogym, graule2021somo, du2021underwater, schaff2022soft, yoo2023toward}. 
Visual monitoring has long been the popular choice for perception in the sim-to-real approach, as it has considerable consistency with the real world. However, the persisting desire for the autonomous operation of soft proprioceptive robots has led to the demand for soft sensor simulation, which in turn suffers from the computational complexity of the robot body. Therefore, the development and maturation of effective sim-to-real technology requires a generalizable, data-efficient sim-to-real adaptation methodology for soft robot proprioception. 

Herein, we propose an unsupervised domain-invariant representation learning approach as a label-free, high-performing, and generalized sim-to-real adaptation method for soft robotic perception. A dual cross-modal autoencoder (AE) enables the alignment of heterogeneous sensor domains at the latent feature level. As a proof of concept, the beneficial features of the proposed framework were examined by applying the framework to a multigait soft robot, which is one of the most popular soft robots, equipped with liquid metal (EGaIn) soft strain sensors. The calibration process was performed for dual principal perception tasks, shape estimation, and collision detection, which are predominantly involved in robotic exploration (Fig. \ref{Main}(b)). An extensive comparative analysis with state-of-the-art methods highlighted the effectiveness of the proposed method in both simulated and real configurations toward accurate digital twinning (Fig. \ref{Main}(a)). 

The principal contributions of the proposed method are as follows:
\newenvironment{packed_enum}{
\begin{enumerate}
  \setlength{\itemsep}{1pt}
  \setlength{\parskip}{0pt}
  \setlength{\parsep}{0pt}
}{\end{enumerate}}

\begin{packed_enum}
    \item Using mapping at the latent level embedded in the unsupervised domain adaptation, our method requires only labeled data from simulations for training, making it intuitively scalable for multitask scenarios.
    \item As our method leverages a dual AE architecture, our model can concurrently perform domain adaptation and anomaly detection. This dual functionality not only facilitates the transfer of tasks trained in the simulation but also aids in collision detection in simulated and real-world domains.
    \item By synergizing accurate kinematic predictions with robust collision detection, our framework enhances the autonomous decision-making capabilities of soft robots. Furthermore, it ensures their safe and efficient operation even in dynamic, unpredictable environments.
\end{packed_enum}

The rest of this article is organized as follows. Section II introduces related works. Subsequently, Section III presents the definition of our target problem. Section IV describes the proprioceptive soft robot and dataset generation strategy adopted in our simulation and experiments. Section V explains the concept of the unsupervised sim-to-real adaptation method, including the baseline methods of domain adaptation and anomaly detection. Section VI presents a quantitative and qualitative performance validation of our framework, and finally, Section VII concludes this article.

\section{Related works}
\subsection{Soft robotics simulation}
The unique, unusual properties of soft robots, such as large and infinite degrees of freedom in deformability, and intricate environmental interactions, demand specialized simulation tools. Early researchers primarily relied on analytical models based on simplified assumptions. Examples of such models include the piecewise constant curvature model \cite{webster2010design} for kinematics and the Cosserat rod model \cite{della2018dynamic} with quasistatic assumptions for dynamics. However, these approximations often cannot capture the intricate nonlinear dynamics of soft robots. 
To mitigate the difficulties in analytical modeling, data-driven models \cite{thuruthel2018model, bruder2020data, bensadoun2022neural} were developed based on empirical robot measurements. However, these models assume the availability of voluminous, representative datasets of soft robots. 
A compelling alternative strategy is to employ models that discretize the continuum structures of the robots and solve the equations of motion for each discretized element. Voxelyze, for example, is a specialized simulation tool developed for soft robots that are voxelizable, and it represents robots as a collection of voxels, each possessing mass and rotational inertia, and connected by beams \cite{hiller2014dynamic}.

For more realistic simulations, finite element methods are the best method to accurately reproduce the dynamics of soft robots \cite{dubied2022sim, PhysX}. Among them, the Simulation Open Framework Architecture (SOFA) \cite{faure2012sofa} is the most widely used tool and offers fast, real-time control for a diverse range of robot geometries and actuators via traditional and reduced-order modeling methods \cite{goury2018fast, katzschmann2019dynamically, tonkens2021soft}.
Based on the wide range of features of SOFA, recent advancements such as SofaGym \cite{schegg2023sofagym} demonstrated sim-to-real transfer in reinforcement learning across various robotic tasks. Currently, this simulator depends on the feedback from images (or, simulator scenes). Meanwhile, simulation with soft proprioceptive sensors has garnered attention from the viewpoint of achieving the autonomous driving of the robots. For instance, the piezoresistive change in liquid-metal-based sensors, which is a popular method to directly measure deformation, was simulated to optimize the design of the sensors within the robot body\cite{tapia2020makesense}. Similarly, Navaro et al. \cite{navarro2020model} modeled the liquid-metal-based capacitive sensors in SOFA, enabling the estimation of the applied force and shape of soft fingers and pads. Although their results showed considerable consistency with the real world, the scope of these works is typically restricted to bending bars fixed at one end, and there exists a gap with the real world restricting the accurate operation of the robot.

\subsection{Unsupervised domain adaptation}
Unsupervised domain adaptation (UDA) is the subfield of domain adaptation that governs knowledge transfer from a source domain to a target domain. The method can serve as a basis for the sim-to-real transfer approach to deal with unpredictable and stochastic uncertainties in the real world. The UDA mainly mitigates the domain shift by obtaining a domain-invariant feature space. UDA methods can be categorized into the following groups: discrepancy-based methods minimize the domain difference using diverse feature distance measurements \cite{long2015learning, sun2016deep}. In contrast, adversarial-based methods adopt the discriminator on the domain classification as generative adversarial networks (GANs) \cite{goodfellow2020generative}, and negative gradient backpropagation is applied to train the domain identification process to make the domains indistinguishable \cite{ganin2016domain}. Meanwhile, reconstruction-based methods obtain feature spaces that are either private in each domain or shared across the domains \cite{bousmalis2016domain}. Although most of these methods were proposed for computer vision applications, in a few studies, they were applied to time-series data \cite{wilson2020multi}. See \cite{wilson2020survey} for a thorough survey of the relevant literature.

\subsection{Time series anomaly detection}
Anomaly detection (AD), i.e., the method of identifying outliers that deviate significantly from expected patterns, is crucial in various practical applications \cite{darban2022deep}. Recently, deep-learning-based approaches for AD have been applied to time-series data. These methods can learn intricate, non-linear patterns and long-term temporal dependencies, thus capturing anomalies that are often overlooked by conventional techniques. Generally, deep-learning-based methods can be classified into two categories: prediction-based and reconstruction-based.
The prediction-based approach \cite{malhotra2015long, ergen2019unsupervised} leverages a trained model to predict a point or subsequence using historical data, designating anomalies based on their deviations from actual values. Although most time-series AD methods rely on prediction errors, it is noteworthy that their efficacy diminishes with rapidly and continuously varying data. As the amount of data increases, the prediction error tends to escalate, constraining these techniques toward short-term predictions to maintain acceptable accuracy. Conversely, reconstruction-based methods \cite{malhotra2016lstm, park2018multimodal} excel with long-term time-dependent data. They focus on the reproduction of data and capture patterns adeptly. The anomalies are identified by gauging the reconstruction errors—the disparities between the reconstructed and original data.

\section{Problem definition}

Soft robots may have to interact dexterously with their surroundings, and they greatly benefit from their morphological compliance. It is interesting to explore the deployment of soft robots in a range of challenging environments that may be unstructured, uneven, and contact-rich. To achieve intelligent servoing in such deployment, two proprioceptive modalities are required: (i) Body shape estimation serves as the basis for providing kinematic information about the robot. Owing to the continuum mechanics of the soft robot, the kinematics is formulated as a distributed form of displacement. (ii) In collision detection, the reactive control strategy secures the robots and their surroundings. Collisions can be detected from abnormal sensor readings with respect to the control input command, which can be interpreted in the noncontacted state. As collision is an unstructured event, the perception should be generalized to various forms of collisions.

From the statements, our simulation-to-real transferable framework performs the aforementioned tasks, both of which are mediated with the UDA strategy. The use of a dual AE for domain adaptation facilitates the driving of the two tasks in a computationally efficient manner; the intermediate latent space promotes domain-adapted shape estimation, and the reconstruction error of the AE output facilitates AD. The training of the dual AE obviates the need for real-world labeling and promotes the effective deployment of various tasks for soft robots.

\section{Proprioceptive multigait soft robot}
\subsection{Soft robot design and fabrication}

\begin{figure}[t]
\begin{center}
\includegraphics[width=\columnwidth] {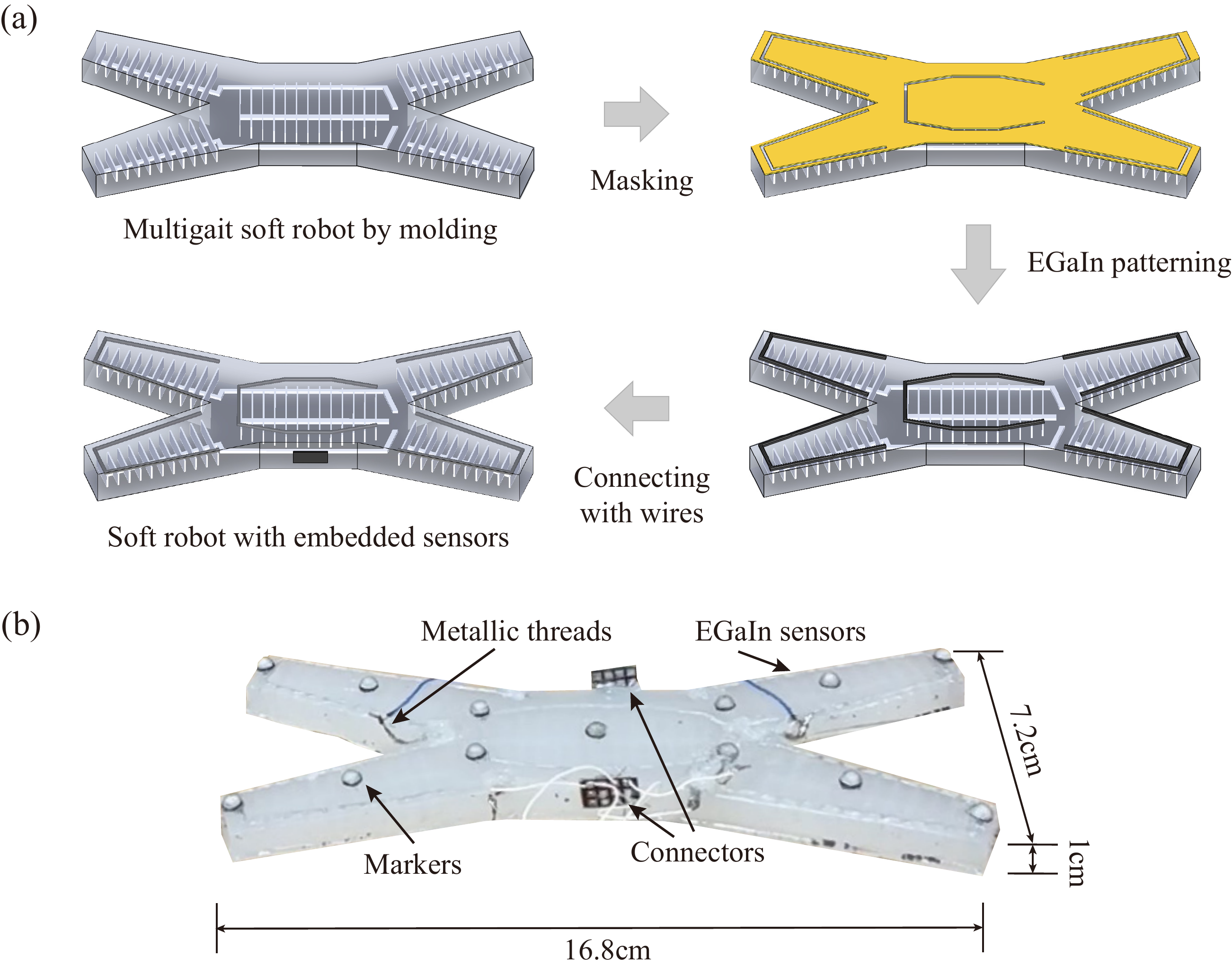}
\caption{\label{FabricationProcess} Fabrication of a proprioceptive multigait soft robot. (a) Craft flow diagram of the strain sensors on multigait soft robot. (b) Top view of the fabricated soft robot and its dimensions.}
\end{center}
\end{figure}
The multigait soft robot consists of five air chambers, each of which governs the bending of four legs and a central body. The bending actuation is performed by a mismatch in the mechanical deformation of a bi-layered structure comprising a stacked inflating chamber containing cavities and a strain-limiting base. The resulting specifications (including stiffness, force, and deformation) can be tuned by appropriately selecting the material for each layer. Herein, we used highly extensible silicone (Ecoflex 00-50; Smooth-On, Inc.) as the chamber and polydimethylsiloxane (Sylgard 184, Dow Corning) as the base. The robot was fabricated by casting on a three-dimensional (3D)-printed mold, followed by bonding of the layers with silicone adhesives. We referred to the previous work \cite{shepherd2011multigait} for the detailed fabrication process.

We then embedded liquid metal (EGaIn) soft strain sensors in each chamber for performing proprioception. These sensors are popular in fields related to soft robotics owing to their high repeatability, fabrication scalability, and adequate deformability \cite{dickey2017stretchable}. The bending deformation of the robot is measured from resistance changes ($\Delta R$) in the liquid metal pathway;  $\propto \Delta l / \Delta A $, where $l$ is the pathway length and $A$ is the cross-sectional area. To prevent electrical disconnection in the liquid metal pathway because of excessive chamber inflation, the EGaIn material is patterned along the edge of the air chamber, as shown in Fig. \ref{FabricationProcess}(a). The fabrication of these sensors first involves printing the EGaIn material on the inflating layer using a stencil mask. An electrical connection is then established at the end of the EGaIn pathway by attaching metallic threads, which offer a reliable connection owing to their high mechanical affinity to liquid metals. The interconnection between electrical wires (UL-AWG24) and the metallic threads is mediated by small electrical boards located on the sides of the robot body. These boards prevent thread detachment from the liquid metal by isolating the wire tension, which is mostly driven by wire pulling. After the wires and threads are connected, the wires are covered with a thin layer of extensible silicone. This step concludes the fabrication process. A graphical introduction to the fabrication process and fabricated soft robot is shown in Fig. \ref{FabricationProcess}.

\subsection{Simulation} 
\begin{figure}[t]
\begin{center}
\includegraphics[width=\columnwidth] {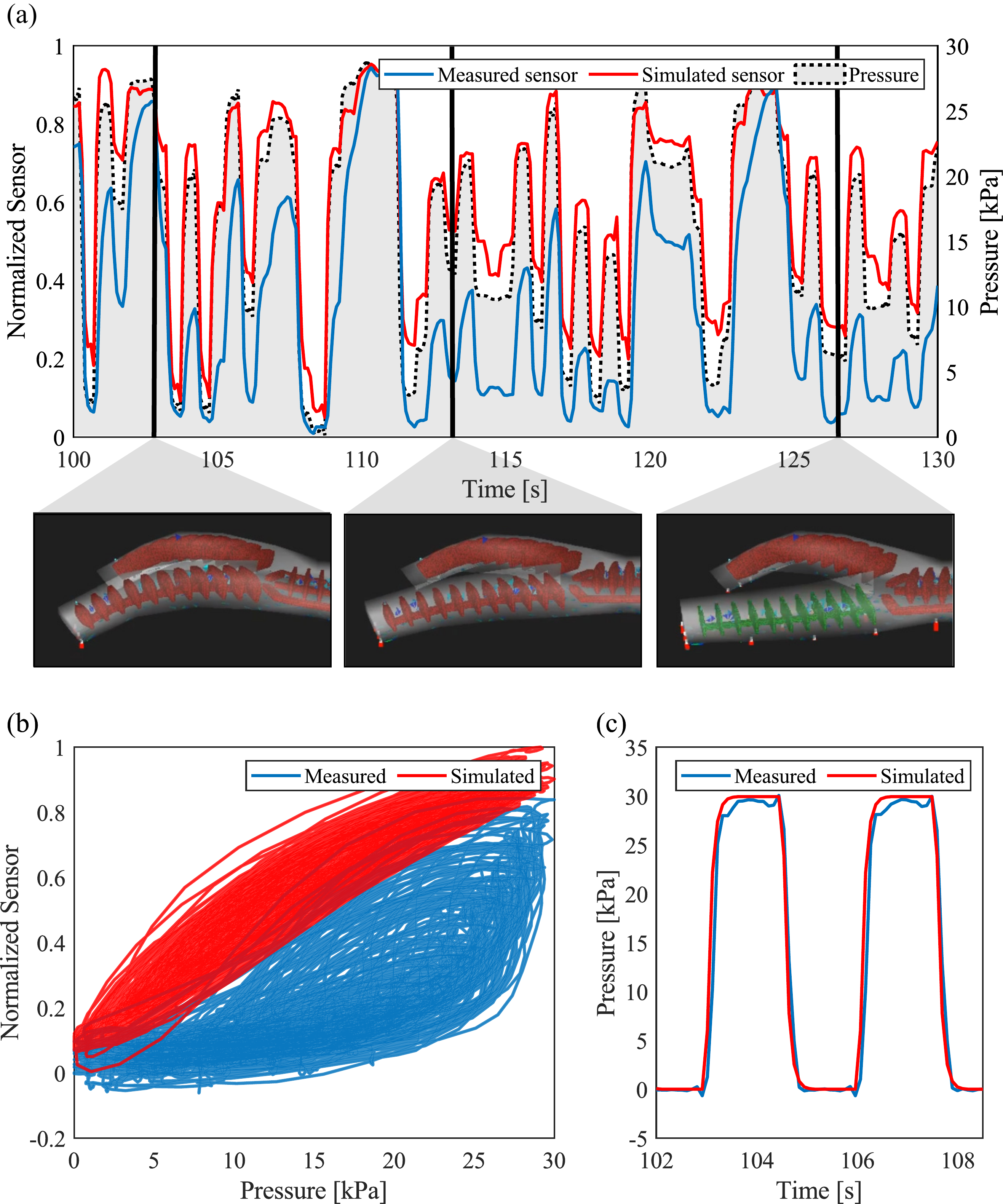}
\caption{\label{Comparison} Comparison of the simulated and the real worlds: (a) rear-right leg sensor signal and its configuration, (b) the hysteresis loop between the normalized sensor signal and pressure, and (c) pressure input.}
\end{center}
\end{figure}
To enable cost-efficient computation, we adopted an open-access reduced-order model of a multigait soft robot based on SOFA \cite{goury2018fast}. 
The modeled robot was constructed with two distinct finite element model components, involving a volumetric part representing the expandable upper layer and a two-dimensional (planar) layer for the inextensible bottom base. The resulting hyper-reduced mesh consists of 2,509 nodes and 8,222 elements, allowing accelerated computation without compromising accuracy.

We simulated the soft sensor behavior by selecting the nodes of the modeled robot along the sensor pathway. The change in resistance at these nodes was calculated based on the sensor geometry. Following Pouillet's law and Poisson's ratio as done in \cite{tapia2020makesense}, the result of the sensor model can be simplified to the following relationship that depends only on sensor length:  
\begin{equation}
R = R_{0}  \left( \left( 1+\frac{\Delta l}{l_{0}}\right)^2-1\right) \label{eq.1}
\end{equation}
where $l$ and $R$ are the sensor length and resistance, respectively, and subscript ${(\cdot)}_0$ indicates the value at rest. 
A comparison of the modeled sensor and actual sensor data is shown in Fig. \ref{Comparison}(a) and (b), clearly revealing their close resemblance in terms of variations.

\subsection{System setup}
\begin{figure}[t]
\begin{center}
\includegraphics[width=\columnwidth] {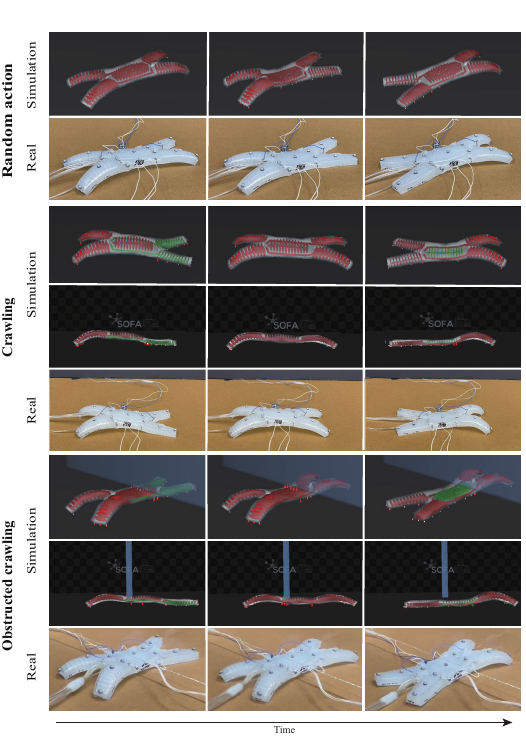}
\caption{\label{Motion1} Shape change for each motion over time of simulated robots and real-world robots.}
\end{center}
\end{figure}
\begin{figure}[t]
\begin{center}
\includegraphics[width=\columnwidth] {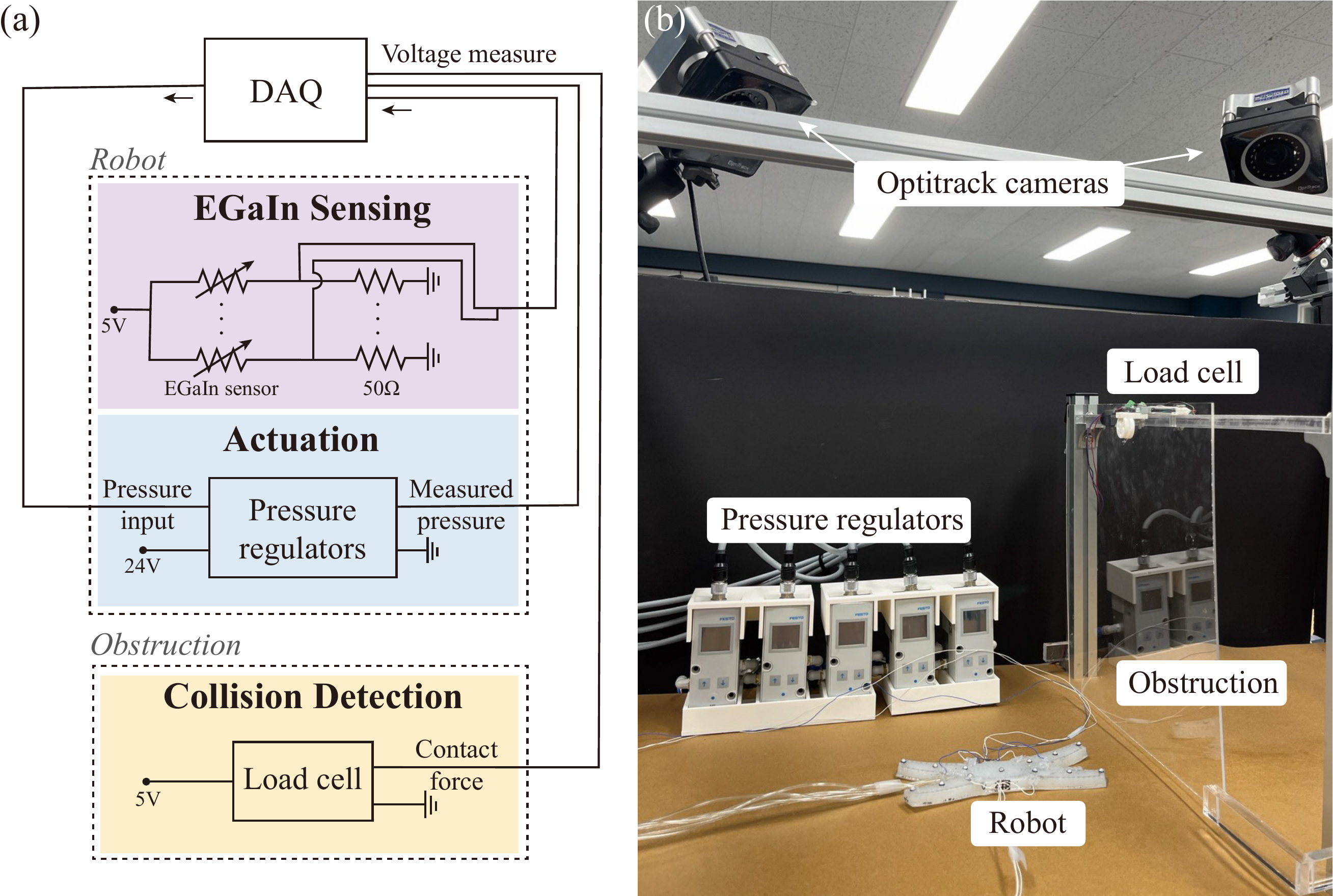}
\caption{\label{System setup} Experimental setup. (a) Circuit block diagram of the control and sensing in the proposed proprioceptive robot with collision detection. (b) System setup with measurements, actuators, and robot. }
\end{center}
\end{figure}

Fig. \ref{System setup} presents the block diagram of the entire circuit and the system setup for the experiment.
The environment for robot operation occupied a workspace of $1.1 \times 1.1 m^2 $ to ensure sufficiently free robot locomotion. At the center, a wall, which served as an obstruction, was installed (Fig. \ref{Motion1}), and space was provided at the bottom to allow the robot to pass. The drive of the robot involved the controlled pressurization of pneumatic channels using five proportional pressure regulators (VPPM, Festo) driven by a microcontroller unit (myRIO-1900, National Instruments). The resulting pressure value was simultaneously measured by the same devices. We set the maximum pressure to 30 and 35 kPa for the legs and body, respectively, to avoid exceeding the strain limit range of the sensors. The sensor resistance value was measured by a voltage-dividing circuit with reference resistance $R$. The robot kinematics was measured by a motion capture system with four OptiTrack cameras (OptiTrack Prime 41, NaturalPoint Inc.). In all, 13 markers were placed on the robot - 12 markers on its legs and one marker on the body - as shown in Fig. \ref{FabricationProcess}(b). The contact of the robot with the wall was measured by a sensitive load cell (SEN-14727, SparkFun Electronics) placed at the top of the wall.

\subsection{Data collection}
The data collection in both the simulated and real worlds was performed by measurements (i.e., the sensor value, pressure, kinematics, and contact force on the wall) under precisely scheduled robot locomotion. Two locomotion styles were embedded in the soft robot operation. In an unobstructed environment, we pressurized the robot with randomly generated pressure inputs to each of the five pneumatic chambers, with a frequency of 0.5s over a duration of 5 min. In addition, crawling motion was realized through seven manually designed steps of actuation sequence performed in steps of 0.5s: (i) starting from the rest state, and then pressurizing the (ii) two rear legs, (iii) central body, and (iv) front legs, and finally, depressurizing (v), (vi), and (vii) them in the same order. All the resulting measurements were performed at 100 Hz and then low-pass filtered and downsampled to 10 Hz for smoothing.

The simulation followed an identical drive protocol as the real world. The frame rate of the simulation was set as 0.02 s to fit the real-world period of measurement. To achieve realistic system control during simulation, the pressure trajectory was emulated the same as that in the real world, as shown in Fig. \ref{Comparison}(b). To be specific, we bounded the rate of the pressure change for each simulation step, and gradually increased the pressure input until it reached the designated target value. As briefly described in Section IV-B, the sensor value was derived from the aforementioned model (\ref{eq.1}). The contact of the robot body with the wall was monitored by computing the reaction force. Lastly, we gathered the 3D positions of 123 nodes to train the kinematic estimation.

\section{Sim-to-real adaptation with a Dual autoencoder}
In the following subsections, we first describe the proposed network of a dual AE. Then, we review the baseline methods for validating our proposed network. Lastly, we explain the details of training and implementation for AD.

\subsection{Proposed network}

\begin{figure}[t]
\begin{center}
\includegraphics[width=\columnwidth] {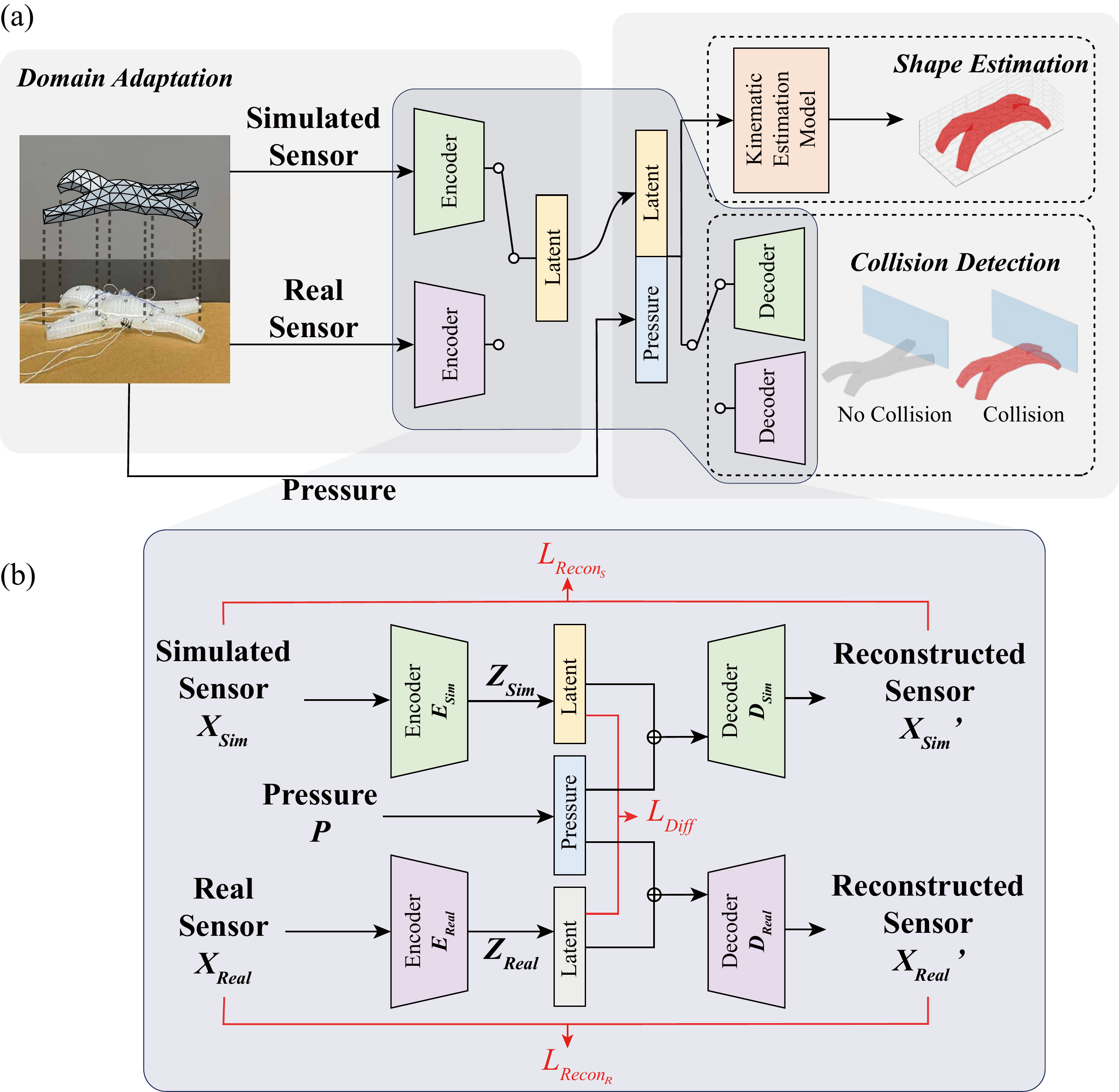}
\caption{\label{Entire structure} Our proposed network of domain adaptation with multiple tasks by using a dual cross-modal autoencoder (AE). (a) Features are extracted from sensor domains and input into the kinematics estimation model or decoder along with pressure. (b) Neural network architecture for domain-invariant latent representation learning.}
\end{center}
\end{figure}

Taking inspiration from the principles of domain adaptation, we aligned domains by matching them within the latent space. There clearly exists a shared feature space between the simulated and real domains, excluding unmodeled dynamics. Therefore, these domains can be bridged by creating a domain-invariant feature representation.

The core of our methodology is a long short-term memory (LSTM)-based dual AE architecture. This dual AE is designed to fulfill two primary objectives: reconstruction-based domain adaptation and AD. This strategic design ensures that the training for feature extraction intrinsically supports collision detection. Building on this foundation, we utilized the extracted features to further train the kinematic estimation.

The comprehensive structure of our network, including the dual AE, a kinematic estimation model, and a collision detection mechanism, is shown in Fig. \ref{Entire structure}(a). In our method, the dual AE is used to extract the shared latent space from each of the two sensor domains $X_{Real}, X_{Sim}\in\mathbb{R}^5$, where the dimension corresponds to the number of air channels in the robot. These are alternately trained in a ratio of 5:1 for two phases, as described below.

In the first phase, the dual AE is trained for feature extraction. Each encoder $E_{(\cdot)}$ maps the strain sensor data $X_{(\cdot)}$ to the latent space $Z_{(\cdot)}\in\mathbb{R}^5$. In parallel, the decoders $D_{(\cdot)}$ are trained to reconstruct the sensor data from the extracted feature that is concatenated with pressure $P\in \mathbb{R}^5$ to ensure a reliable reconstruction by conditioning to the control variable. Notably, this training process is performed using solely data of the scenario without an obstruction (hereinafter, unobstructed data) to ensure large reconstruction errors when the model is inferred with data of the scenario with an obstruction (hereinafter obstructed data), thereby facilitating effective collision detection. As shown in Fig. \ref{Entire structure}(b), the loss function in this domain adaptation phase, $\mathcal{L}_{DA}$, can be written as

\begin{equation}
\mathcal{L}_{DA} = \mathcal{L}_{Recon_{R}} + \mathcal{L}_{Recon_{S}} + \mathcal{L}_{Diff},\\ 
\label{DA loss} 
\end{equation}

where $\mathcal{L}_{Recon_{(\cdot)}}$ are the reconstruction losses, and $\mathcal{L}_{Diff}$ is the difference loss that estimates the error between the latent variables from the simulated and real domains. These losses can be expressed as

\begin{equation}
\mathcal{L}_{Recon_{(\cdot)}}  = \mathcal{||} X_{(\cdot)} - D_{(\cdot)}(E_{(\cdot)}(X_{(\cdot)}) \oplus P) \mathcal{||}_{2}\\ 
\label{Recon loss} 
\end{equation}

\begin{equation}
\mathcal{L}_{Diff}  = \mathcal{||} E_{Sim}(X_{Sim}) - E_{Real}(X_{Real}) \mathcal{||}_{2}
\label{Diff loss} 
\end{equation}

where $\oplus$ denotes the concatenation of the tensors.

\begin{algorithm}[t]
 \caption{Model training}
 \begin{algorithmic}[1]
  \renewcommand{\algorithmicrequire}{\textbf{Input:}}
   \REQUIRE Sensor data in the simulated domain $X_{Sim}$ and the real domain $X_{Real}$; Kinematics data in the simulated domain $k$; Neural network of each model $\theta_{(.)}$;
 \renewcommand{\algorithmicrequire}{\textbf{Parameter:}}
  \WHILE{ Algorithm Not converge}
  \FOR {i = 1, ..., 5}
  \STATE Compute $\mathcal{L}_{DA}$ from $X_{Sim}$, $X_{Real}$ in the unobstructed scenes using Eq. \eqref{DA loss}
  \STATE Update $\theta_{E_{Sim}}$, $\theta_{D_{Sim}}$,$\theta_{E_{Real}}$, and $\theta_{D_{Real}}$
  \ENDFOR
  \FOR {i = 1} 
  \STATE Compute $\mathcal{L}_{kine}$ from $X_{Sim}$ using Eq. \eqref{Kine loss}
  \STATE Update $\theta_{E_{Sim}}$ and $\theta_{K}$
  \ENDFOR
  \ENDWHILE
 \end{algorithmic}
 \end{algorithm}

In the second phase, task-specific calibration was performed in the shared latent space. In our two tasks, kinematics estimation was learned through a neural network architecture-- this learning model is referred to as the kinematic estimation model. In line with this task learning, the encoder of the simulated domain $E_{Sim}$ was simultaneously trained to lead to a latent space better favorable to the tasks. As done in the first phase, the tuned features from the simulated sensor data were concatenated with the pressure and served as an input to the kinematic estimation model. However, unlike the first phase, the training in the second phase was guided to experience the behaviors in both obstructed and unobstructed environments. The resulting training minimized the kinematic estimation loss $\mathcal{L}_{kine}$, which is defined as follows:
\begin{equation}
\mathcal{L}_{kine}  =  \mathcal{||} k_t - K(E_{Sim}(X_{Sim}) \oplus p)\mathcal{||} \\
\label{Kine loss} 
\end{equation}
where $k_t$ is the ground truth value, and $K$ denotes the kinematic estimation model. Algorithm 1 summarizes the overall training procedure.

As a neural network architecture, we used the LSTM \cite{hochreiter1997long} to address the time-dependent regimes, which mainly originated from viscoelastic hysteresis in soft body dynamics. Both encoders and decoders had two LSTM layers, each with 256 hidden dimensions. The decoders were further augmented with a single fully connected (FC) layer, designed to output five dimensions, and a rectified linear unit (ReLU) activation function. The kinematic estimation model was designed with an LSTM layer followed by two FC layers with 256 dimensions, sequentially incorporating the ReLU and tanh activation functions.

\subsection{Baseline methods} 
This section describes various baseline methods used to validate the beneficial features in our framework, the performance of domain adaptation, and the AD of our network.

\textbf{Domain Adaptation}

\textit {1) Vanilla LSTM.}
Supervised learning based on the LSTM (vanilla LSTM) has been used in the soft robotics field owing to its ease of handling long-time-lagged data \cite{thuruthel2019soft}. The supervised learning in our work was performed in real-to-sim and sim-to-task mappings, both trained in an end-to-end manner. The model trained in the real-to-sim mapping serves to transfer the real data to the sim-to-task map that is pretrained in the simulation environment. In the following sections, we refer to the model trained only for the sim-to-task transition as sim-only vanilla LSTM and the model trained for the real-to-sim-to-task transition as real-to-sim vanilla LSTM.

\textit {2,3) CoDATS and R-DANN.}
These methods are the most commonly used domain adversarial neural network (DANN)-based approaches for time-series data. Convolutional deep domain adaptation model for time-series data (CoDATS) \cite{wilson2020multi} comprises a set of one-dimensional convolutional neural networks (1D-CNNs), allowing the model to capture spatial dependencies upon temporal evolution. In contrast, recurrent DANN (R-DANN) uses LSTM layers, which can capture long-term dependencies. These models enable us to evaluate the performance of the adversarial approach against the reconstruction-based adaptation of our proposed model.

\textit {4) DSN.}
Domain separation network (DSN) \cite{bousmalis2016domain} was selected as a comparative benchmark because of its close resemblance to our proposed reconstruction-based approach. The challenge of identifying shared representations between domains is the vulnerability to noise contamination that is correlated with the underlying shared distribution. To address this problem, DSN uses a private encoder for each domain to capture domain-specific properties, alongside a shared subspace that is designed to be orthogonal to these private subspaces. Through the deliberate exclusion of the domain-specific properties from the shared subspaces, the DSN can effectively minimize domain discrepancies within this shared subspace. A comparison of the performance of our proposed model with that of the DSN can provide valuable insights into the benefits and limitations of explicitly modeling shared and domain-specific features.

\textit {5) Single AE.}
In a single AE approach, one AE is used, leveraging the same network for feature extraction across each domain. Numerous domain adaptation techniques are built upon deep architectures, wherein the weights are shared for both domains \cite{tzeng2014deep, girshick2014rich, long2015learning, ganin2015unsupervised}. Therefore, we compared our dual AE to a single AE to assess the effectiveness of our domain-specific AE methodology.

\textbf{Anomaly detection}

\textit {Prediction-based 1) real-to-sim vanilla LSTM and 2) dual AE.}
In prediction-based AD for time-series data, LSTM layers are widely employed \cite{malhotra2015long}. To assess the suitability of the reconstruction-based AD of our model for the target sensor domains, we performed prediction-based approaches, training the model to predict either the simulated sensor values or the latent features extracted by our model. For clarity, each method is further denoted as prediction-based real-to-sim vanilla LSTM or prediction-based Dual AE.

\textit {3) Reconstruction-based AD using DSN.}
DSN has also been applied for AD combined with domain adaptation, especially for visual images \cite{yang2023anomaly}. To extend its application to sequential multivariate time-series data, we incorporated an LSTM-based AE \cite{malhotra2016lstm} into the DSN, given the prevalence of LSTM layers in time-series analysis.

Details on the implementation of each baseline method are available in the appendix. The kinematic estimation models were structured identically to that of the proposed method.

\subsection{Training details} 
For the training process, five observations of random actions in both the simulated and real worlds and ten observations of obstructed crawling in the simulated setting were conducted. From these observations, one sample of each observation set was used for validation, while others were used for training. To ensure stable and efficient training, all training data were normalized. 
The models were trained with the Adam optimizer \cite{kingma2014adam} with an initial learning rate of $4\times{10}^{-4}$ for the domain adaptation model and ${10}^{-3}$ for the shape estimation model. Training was performed until validation loss failed to converge, exhibiting monotonous increments for 100 epochs. A weight decay coefficient of ${10}^{-6}$ was applied to all the models to prevent overfitting and encourage generalization.
For a fair comparison, we conducted training five times with random seeds.

\subsection{Implementation of collision detection} 
We conducted ground truth collision labeling based on the contact force, classifying events as collisions only in instances of strong impacts. This approach is grounded in the observation that the deformation of the robot becomes abnormal— i.e., less than usual—exclusively during strong collisions. In line with these characteristics, we used two approaches for error calculation based on the model type. For reconstruction-based methods, we calculated errors only when the reconstructed sensor data exceeded the input sensor data.  In contrast, for prediction-based methods, we used the absolute value of the error to identify collisions.

To establish the threshold for collision detection, we first summed the errors observed across all five sensor channels into a single time series sequence. We then calculated the average of the 10th highest error values from each type of motion—namely, random action and obstructed crawling. The final threshold was determined by obtaining the mean value from five independently trained models, thereby ensuring a more generalized result. 

Given that pressure was applied at intervals of 0.5 s, we segmented the data into sets corresponding to five time steps. Collision labeling was then implemented based on the number of data points that exceeded the predetermined threshold within each of these segments. In the simulation environment, we set this threshold number to 1, while in the real-world experimental setup, we adjusted it to 2 to consider potential noise.

\section{Results}

In this section, we present the results of our experiments, conducted in both the simulated and real-world environments. We first assessed the adaptability across sensor domains through kinematic estimation, providing a comparative analysis with baseline methods. Next, we demonstrated the effectiveness of our approach in collision detection through AD experiments. To ensure the reliability and applicability of our results, we averaged them across five distinct trained models. The performance was evaluated by comparison with the aforementioned baseline methods and with a specific variant of our method--a dual AE without pressure for the decoder, denoted as dual AE (w/out P)). This comparison allowed us to investigate the effects of integrating pressure data into the decoder. We concluded our analysis by explicitly illustrating the reduction of the domain gap, by comparing the latent vectors extracted from each sensor domain.

\subsection{Shape estimation results}

\begin{table*}[tbh]
\caption[Shape estimation Results of the simulated sensor]
{Shape estimation results of the simulated and physical sensors ($\times 10^{-5}$)}
\label{tab:ShapeestimationResults1} 
\begin{tabular*}{\textwidth}{@{\extracolsep{\fill}} c|c|cccccc|cc}
\hline
&&\multicolumn{6}{c|}{Baseline methods}&\multicolumn{2}{c}{Proposed method}\\ 
Domain& Task & Sim-Only & Real-to-sim & \multicolumn{4}{c|}{} &\multicolumn{2}{c}{}  \\ 
&&Vanilla LSTM &Vanilla LSTM & CoDATS & R-DANN & DSN & Single AE &Dual AE (w/out P)&Dual AE \\
\hline
\multirow{3}{*}{Simulation} & Random action & \textbf{624$\pm$43}&-&978$\pm$172&778$\pm$190&745$\pm$65&710$\pm$65&747$\pm$185&780$\pm$34\\ \cline{2-10}

& Obstructed &\multirow{2}{*}{\textbf{230$\pm$15}}&\multirow{2}{*}{-}&\multirow{2}{*}{1141$\pm$129}&\multirow{2}{*}{1227$\pm$153}&\multirow{2}{*}{796$\pm$25}&\multirow{2}{*}{289$\pm$42}&\multirow{2}{*}{295$\pm$88}&\multirow{2}{*}{324$\pm$59}\\
&Crawling&&&&&&&&\\
\hline
Real &Random action& 2390$\pm$183& \textbf{631$\pm$41}& 1315$\pm$454& 843$\pm$163& 745$\pm$65&713$\pm$63&774$\pm$176&782$\pm$36\\
\hline
\end{tabular*}
\end{table*}

\begin{table*}[t]
\caption[Shape estimation Results of the real sensor]
{Error between the shapes estimated from the real sensor and the marker positions of the physical robot [mm]}
\label{tab:ShapeestimationResults2} 
\begin{tabular*}{\textwidth}{@{\extracolsep{\fill}} c|cccccc|cc}
\hline
&\multicolumn{6}{c|}{Baseline methods}&\multicolumn{2}{c}{Proposed method}\\ 

Task& Sim-Only&Real-to-sim & \multicolumn{4}{c|}{} &&\\
&Vanilla LSTM & Vanilla LSTM &CoDATS & R-DANN & DSN & Single AE &Dual AE (w/out P)&Dual AE \\
\hline
Random action& 2.12$\pm$0.082&1.69$\pm$0.020& 1.81$\pm$0.182&1.79$\pm$0.061& 1.70$\pm$0.014& 1.70$\pm$0.004& \textbf{1.65$\pm$0.015 }&1.68$\pm$0.001\\
\hline
Obstructed &\multirow{2}{*}{2.14$\pm$0.193}&\multirow{2}{*}{1.91$\pm$0.123}&\multirow{2}{*}{1.99$\pm$0.067}&\multirow{2}{*}{2.03$\pm$0.098}&\multirow{2}{*}{1.92$\pm$0.004}&\multirow{2}{*}{1.92$\pm$0.011}&\multirow{2}{*}{\textbf{1.81$\pm$0.009}}&\multirow{2}{*}{1.87$\pm$0.049}\\
Crawling&&&&&&&&\\
\hline
\end{tabular*}
\end{table*}

\begin{figure}[t]
\begin{center}
\includegraphics[width=\columnwidth] {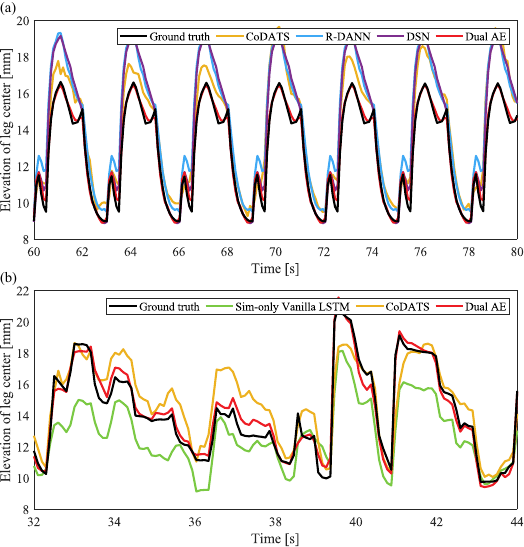}
\caption{\label{Comparison_plot}  Comparison of kinematics estimation (a) in the simulation domain during obstructed crawling, and (b) in the real domain during random actions.}
\end{center}
\end{figure}

\begin{figure*}[t]
    \centering
    \resizebox{\textwidth}{!}{\includegraphics{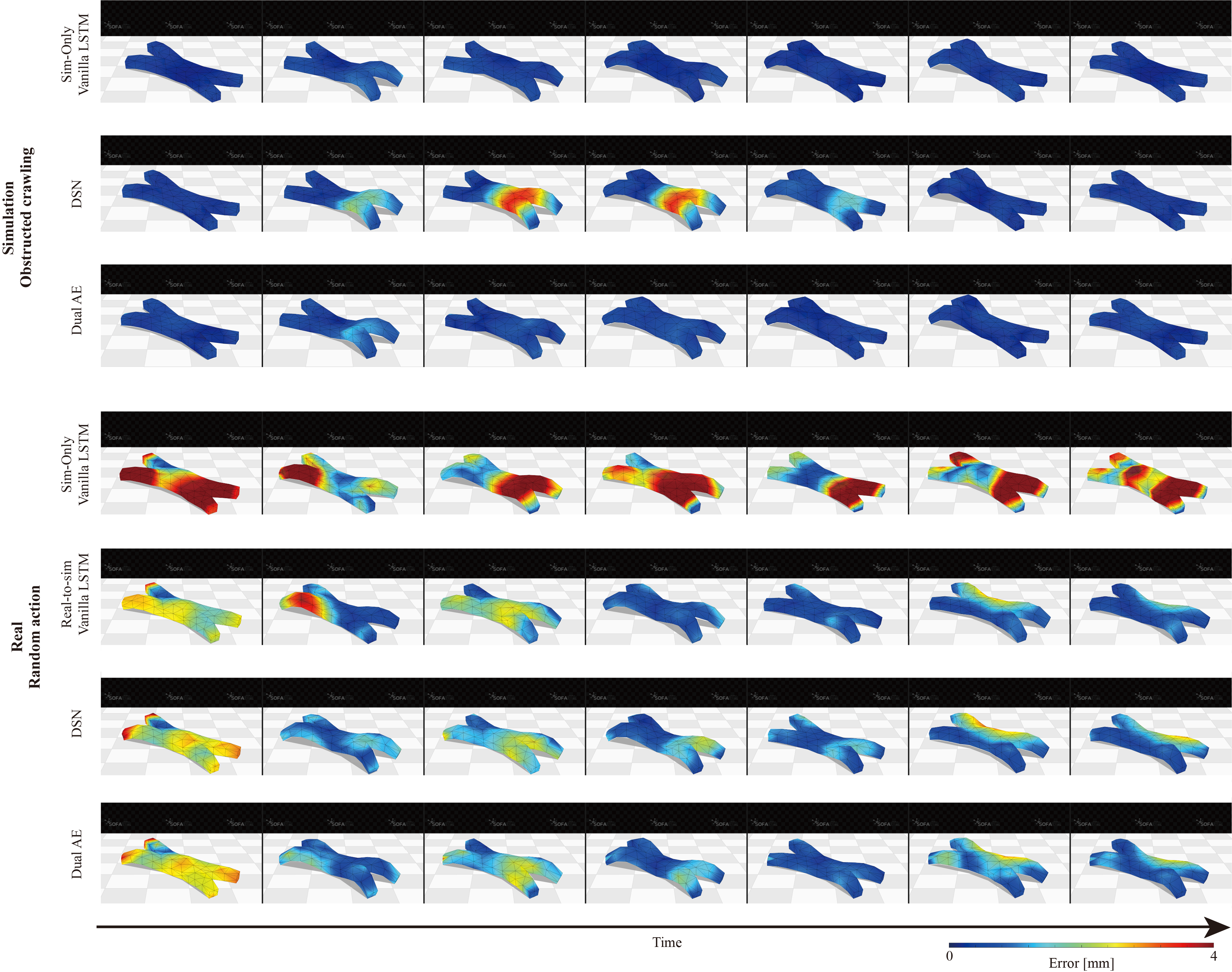}}
    \caption{\label{SE_result} Comparison of the estimated kinematics results with the error for each adaptation method. The upper rows show the results from the simulation domain for obstructed crawling, whereas slower rows show the results from the real domain for random action.}
\end{figure*}

We evaluated the performance of domain adaptation in each domain for two scenarios: random action and obstructed crawling. 
Table \ref{tab:ShapeestimationResults1} summarizes the results of shape estimation based on physical and simulated sensors. To assess the accuracy of shape estimation, we computed the mean absolute error for the XYZ positions of nodes over 3000 time steps.

\subsubsection{Simulation domain}
In the simulation domain, all adaptation methods successfully estimated shapes during random actions. However, for obstructed crawling, the adversarially trained models (i.e., CoDATS and R-DANN) and DSN had larger errors. This result implies that the features extracted by these models fail to represent sensor dynamics. Fig. \ref{Comparison_plot}(a) supports the aforementioned performance comparison in detail.

\subsubsection{Real domain}
While all methods effectively estimated shapes for random actions in the simulated sensor data, differences emerged when using physical sensor data. From Fig. \ref{Comparison_plot}(b) and Table \ref{tab:ShapeestimationResults1}, we see that vanilla LSTM struggles to accurately estimate kinematics without real-to-sim adaptation, highlighting the necessity for domain adaptation. Furthermore, the performance of the adaptation of adversarially trained models is limited, and R-DANN performed better than CoDATS in the real and simulated domains. This result indicates that compared to CNN layers, LSTM layers are better suited for soft robots that have high hysteresis and require long-term memory. Interestingly, DSN produces consistent errors in both the simulated and real domains, implying that its kinematic model relies only on pressure data, as also seen in Fig. \ref{Comparison_plot}(a). The errors in our methods closely align with those in the simulation domain and the real-to-sim vanilla LSTM, confirming their adaptability.

In the case of obstructed crawling, generating equivalent simulated shape labels for both simulation and real-world scenarios was not feasible. Instead, we used marker positions obtained via motion tracking to calculate the error between the estimated and actual shapes. Specifically, we selected five markers located in the middle of each chamber where the largest deformation changes occurred. The height difference between these markers and their corresponding nodes in the finite element model was then computed.
The results in Table \ref{tab:ShapeestimationResults2} indicate that the error trends obtained using actual shape labels closely mirror those from the simulated shape labels in random action scenarios. However, when it comes to obstructed crawling scenarios, the errors increase. This result implies that although shape estimation is feasible, the model accuracy in real-world obstructed crawling scenarios is less than that in the simulations, as these conditions are unseen to the kinematics model. The dual AE method consistently showed the lowest error rates, while the adversarial methods performed less effectively, similar to their performance under simulated conditions.

Fig. \ref{SE_result} offers a visual comparison of the estimated shapes with their errors for obstructed crawling in the simulation domain and random action in the real domain. Well-performing baseline methods, i.e., DSN and real-to-sim vanilla LSTM, are shown along with our proposed model for a comprehensive comparison.
An elaborate comparison of the marker data and estimated shape of the physical robot across all methods can be seen in the supplementary video.

\subsection{Collision detection results}

\begin{figure*}[t]
    \centering
    \resizebox{\textwidth}{!}{\includegraphics{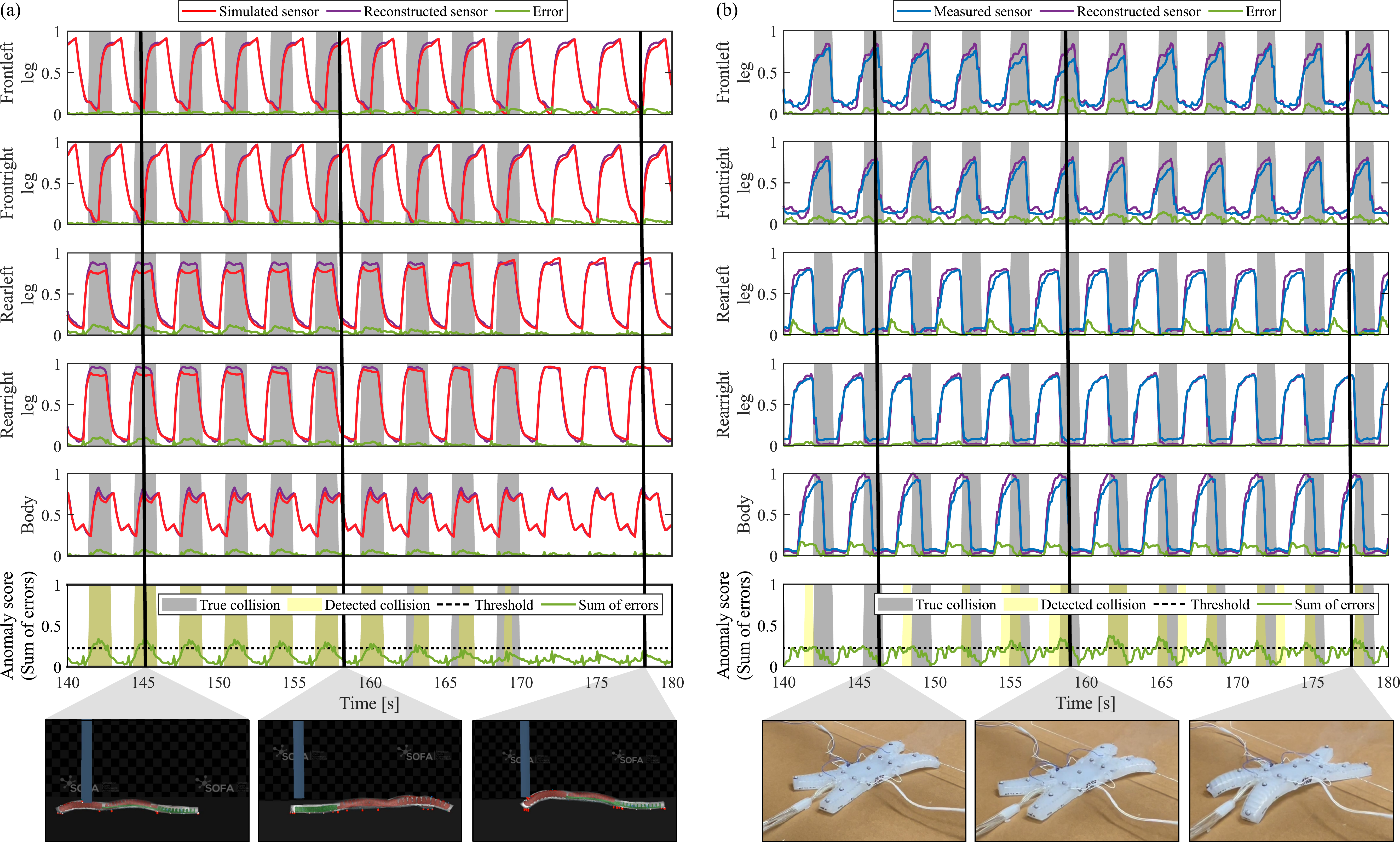}}
    \caption{\label{AD_sensor} Collision detection results of our proposed model (dual AE) with the comparison of the input sensor and reconstructed sensor signals change over time, and the robot configuration at the time spots in (a) the simulation and (b) real world. The data are shown after normalization.}
\end{figure*}

\begin{figure}[t]
\begin{center}
\includegraphics[width=\columnwidth] {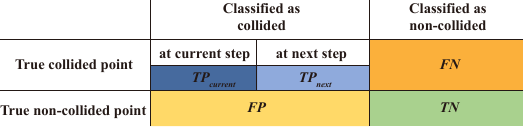}
\caption{\label{confusion mat} Confusion matrix used in our work. TP, FN, FP, and TN indicate true positive, false negative, false positive, and true negative, respectively.}
\end{center}
\end{figure}

\begin{figure}[t]
    \centering
    \resizebox{\columnwidth}{!}{\includegraphics{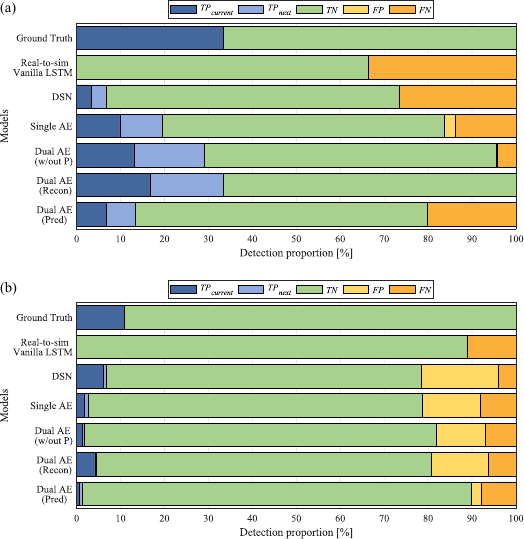}}
    \caption{\label{AD_result} Collision detection results of obstructed crawling in the (a) simulation and (b) the real world.}
\end{figure}

\begin{table*}[t]
\caption[The accuracy of the collision detection]
{Accuracy and F1 scores of the collision detection}
\label{tab: Collision detection result} 
\begin{tabular*}{\textwidth}{@{\extracolsep{\fill}} cc|c|ccc|ccc}
\hline
\multicolumn{1}{c|}{\multirow{4}{*}{Domain}} &  \multirow{4}{*}{Task} & \multirow{4}{*}{Metric}  &  \multicolumn{3}{c|}{Baseline methods} &  \multicolumn{3}{c}{Proposed method} \\
\multicolumn{1}{c|}{} &   &   &  \multirow{2}{*}{\begin{tabular}[c]{@{}c@{}}Real-to-sim\\ 
Vanilla LSTM\end{tabular}} &  \multirow{2}{*}{DSN} &  \multirow{2}{*}{Single AE} &  \multirow{2}{*}{Dual AE (w/out P)} &  \multirow{2}{*}{Dual AE} &  \multirow{2}{*}{Dual AE} \\
\multicolumn{1}{c|}{} &   &   &   &   &   &   &   &   \\ \cline{4-9} 
\multicolumn{1}{c|}{} &   &   &  Prediction &  Reconstruction &  Reconstruction &  Reconstruction &  Reconstruction &  Prediction \\ \hline
\multicolumn{1}{c|}{\multirow{5}{*}{Simulation}} &  \multirow{1}{*}{Random action} &  A &  1 &   1 &  1 &  1 &  1 &  0.998 \\ \cline{2-9} 
\multicolumn{1}{c|}{} &  \multirow{4}{*}{\begin{tabular}[c]{@{}c@{}}Obstructed\\ Crawling\end{tabular}} &  P &   - &   1&   0.894&   0.997&   1&   1\\ \cline{3-9} 
\multicolumn{1}{c|}{} &   &  R &   0 &   0.202&   0.584&   0.872&   1&   0.4\\ \cline{3-9} 
\multicolumn{1}{c|}{} &   &  F1 &   - &   0.336&   0.706&   0.930&   1&   0.571\\ \cline{3-9} 
\multicolumn{1}{c|}{} &   &  A &  0.665 &  0.734 &  0.838 &  0.956 &  1 &  0.799 \\ \hline
\multicolumn{1}{c|}{\multirow{5}{*}{Real}} &  \multirow{1}{*}{Random action} &  A &  1 &  0.999 &  1 &  0.990 &  1 &  0.952 \\ \cline{2-9} 
\multicolumn{1}{c|}{} &  \multirow{4}{*}{\begin{tabular}[c]{@{}c@{}}Obstructed\\ Crawling\end{tabular}} &  P &   - &   0.279&   0.169&   0.142&   0.259&   0.134\\ \cline{3-9} 
\multicolumn{1}{c|}{} &   &  R &   0 &   0.629&   0.247&   0.210&   0.423&   0.346\\ \cline{3-9} 
\multicolumn{1}{c|}{} &   &  F1 &- &0.387&0.201&0.169&0.321&0.193\\ \cline{3-9} 
\multicolumn{1}{c|}{} &   &  A &  0.889 &  0.784 &  0.787 &  0.819 &  0.807 &  0.899 \\ \hline
\multicolumn{2}{c|}{\multirow{4}{*}{Mean}} &  P &
   - &   \textbf{0.640}&   0.532&   0.570&   0.630&   0.567\\ \cline{3-9} 
\multicolumn{2}{c|}{} &  R &   0 &   0.416&   0.416&   0.541&   \textbf{0.712}&   0.373\\ \cline{3-9} 
\multicolumn{2}{c|}{} &  F1 &   - &   0.504&   0.466&   0.555&   \textbf{0.668}&   0.450\\ \cline{3-9} 
\multicolumn{2}{c|}{} &  A &  0.889 &  0.879 &  0.906 &  0.941 &  \textbf{0.952} &  0.912 \\ \hline
\end{tabular*}
\end{table*}

Fig. \ref{AD_sensor} displays sensor data and its errors with detection results based on our method. The figure shows that an obstruction reduces the input sensor signal. Fig. \ref{confusion mat} presents the confusion matrix used for evaluating the performance of our method. We recorded true positives (TPs) when anomalies were detected either in the next time step or in the current step following a true anomaly. This is to reflect the inherent time delays in soft robot reactions. For a more detailed analysis, we separately record TP at the next and current steps. AD results for obstructed crawling in both real and simulated domains are visualized in Fig. \ref{AD_result}.
Based on these criteria, we evaluated the results using F1-score $F1$ and accuracy $A$, which are defined as follows:

\begin{equation}
F1  = \cfrac{2 \cdot P \cdot R}{(P + R)},  P  = \cfrac{(TP)}{(TP + FP)},  R  = \cfrac{(TP)}{(TP + FN)}
\label{eq:F1score}
\end{equation}

\begin{equation}
A  = \cfrac{( TP + TN ) }{(TP + TN + FP + FN)} 
\label{eq:accuracy}
\end{equation}

where $TP$ indicates $TP_{next} + TP_{current}$ and $P$ and $R$ denote precision and recall, respectively. $TN, FP$, and $FN$ refer to true negative, false positive, and false negative, respectively.

A higher accuracy and higher F1 score indicate correct data point classification. While accuracy provides an overall measure of detection correctness, the F1 score is more appropriate for imbalanced datasets. In general, the F1 score is commonly used as an evaluation metric of AD, but for random action scenarios where TP values are not available, only accuracy is used to evaluate performance. For evaluations in obstructed crawling, we calculated both metrics to comprehensively evaluate the model performance. The results are summarized in Table \ref{tab: Collision detection result}. 

\subsubsection{Simulation domain}
In random action scenarios, all methods except for the prediction-based ones performed well. However, in obstructed crawling, only the reconstruction-based dual AE achieved an accuracy and F1 score of 1, while single AE and dual AE (w/out P) showed good results in shape estimation. Detailed results are shown in Fig. \ref{AD_result}(a), where FNs are prevalent in other methods. This is because the decoders in the other methods can reconstruct even abnormal sensor data similarly, resulting in small reconstruction errors.

\subsubsection{Real domain}
While prediction-based methods achieve the highest accuracy, Table \ref{tab: Collision detection result} summarizes their limitations in collision detection, leading to low F1 scores. Notably, DSN and dual AE stand out with the highest F1 scores. However, it is important to note that all methods, including these two, exhibit high FP rates in real-world scenarios, which diverges from their performance in simulations and leads to overall low accuracy.
This deviation can be attributed to the inherent dynamics of the physical robot, such as less-pronounced initial deformations and delayed sensor responses to obstructions. These factors contribute to large reconstruction errors, which in turn lead to mislabeling events as collisions. In contrast to its simulation performance, dual AE shows elevated FN values because of the denoising process applied to real sensor data during latent matching. Therefore, smaller sensor variations are often smoothed out in the reconstructed data and remain undetected, leading to an increase in FN values.

Among all scenarios in both domains, it is evident that dual AE consistently shows the highest average of accuracy and F1 score. These results demonstrate the effectiveness of the reconstruction-based dual AE for detecting collisions in both simulated and real sensor domains compared to the other AD methods.
Moreover, although dual AE (w/out P) showed slightly better results (by 2.57\%) in shape estimation, it lagged significantly in collision detection, with a performance disparity of 10.76\%. This differential highlights the advantages of integrating pressure into the decoder when performing domain adaptation and collision detection tasks.

\subsection{Model Analysis}

\begin{figure}[t]
\centering
\resizebox{\columnwidth}{!}{\includegraphics{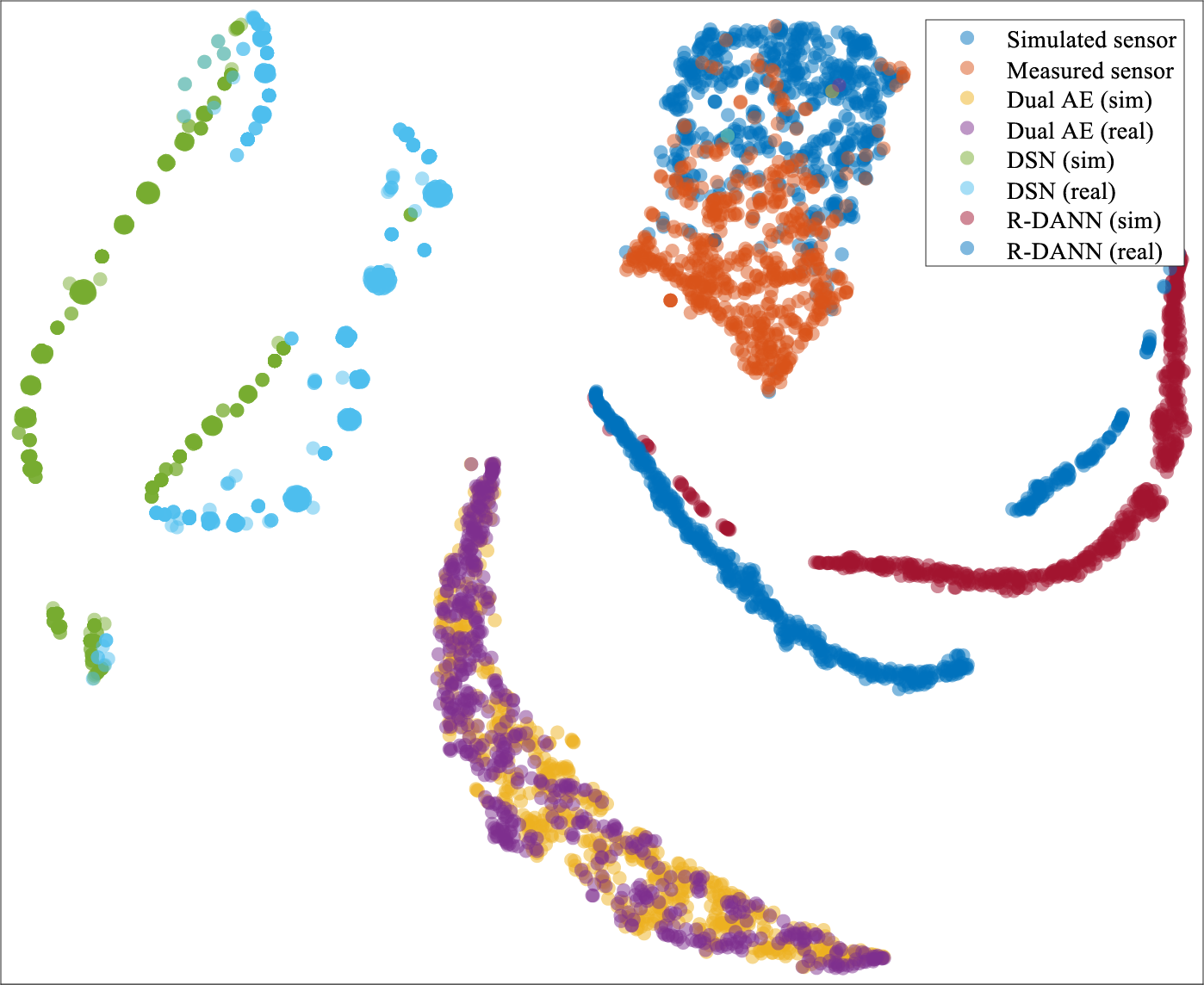}}
\caption{\label{tSNE} Data distributions of sensor data and extracted features from the simulated and real domains.}
\end{figure}

To identify the reduction of the gap between domains using each method, we used the t-distributed stochastic neighbor embedding (t-SNE) \cite{van2008visualizing} to compare the extracted features from each domain. By projecting them from a high-dimensional space to a two-dimensional (2D) plane using t-SNE, we can visualize and compare the characteristics of these two feature vectors. In this 2D plane, the distance between any two points indicates the similarity of the features that they represent.

As shown in Fig. \ref{tSNE}, the feature clusters from both the simulated and real domains, which represent the sensor data before domain adaptation, share a similar region but with a noticeable gap. Additionally, the flow of features within the feature vector appears disorganized, and this is likely a consequence of the nonlinear attributes of the soft sensors. In contrast, after domain adaptation, the extracted features appear linear, which is characteristic of time-series data, suggesting that the adaptation methods successfully captured the time-series property of the data. However, notably, the DSN and R-DANN methods yield more distinct feature clusters. In comparison, our dual AE model shows regions where the simulated and real domains nearly overlap, indicating the proficient adaptation of the sensors in the two domains. Moreover, the distribution exhibits a more pronounced differentiation than before, underscoring the capability of our model to not only bridge the domain gap but also refine the sensor data for enhanced distinction.

\section{Conclusion}

In this work, we introduced the UDA methodology for sim-to-real bridging of soft robot perception using dual cross-modal AE. The sensor dynamics in these heterogeneous domains are matched at the latent level, eliminating the different properties originating from both domains. Through extensive investigations, we demonstrated the effectiveness of our method compared with previously developed methods in multiple tasks that are crucial and challenging in autonomous soft robot operation. 
Our results show that our framework not only shows comparable performance with supervised learning in domain adaptation but even outperforms it, especially under unseen real-world conditions such as obstructed crawling. This result emphasizes the robustness and generalizability of our latent matching approach.

As mentioned in the Introduction, machine learning provides a means to address the complexity in the modeling of soft robots. Although the approach is of interest in the field of generic robotics, the unique characteristics of soft robots strongly require such data-driven computation, rather than analytical and numerical formulations. For instance, under ideal conditions, highly accurate soft body simulation can achieve computationally efficient calibration process without any domain adaptation process. However, there are challenges posed by variance in the manufacturing process and high complexity (or, often unavailability) in soft continuum mechanics that exhibit nonlinear and contact-rich characteristics. Our simulation achieves adequate computation performance, indicating that our approach is generalizable to various soft robot designs that involve comprehensive actuation mechanisms, geometry, and perception methods and is applicable to many other perception tasks such as terrain classification, and environmental recordings. 

Although we demonstrated a methodology for sim-to-real transferring of sensors via the proprioceptive multigait soft robot, some limitations still remain. First, during the latent matching process, our framework denoizes real-world sensor data. Therefore, any abnormal sensor changes are less reflected in the reconstructed data, leading to higher reconstruction errors compared with the simulations. To alleviate this problem, we summed the errors across all five channels for collision detection, albeit with low sensing resolution and sensitivity. In future research on reducing sensor noise during fabrication or measurement stages, collision detection accuracy can be improved and segment-specific detection may become possible.
In addition, our experiment setup primarily focused on a crawling gait pattern with an obstruction, i.e., a wall. This framework can be extended to various control tasks and obstructions. By incorporating higher-resolution sensors, such as those for the whole-body sensing approach \cite{park2021deep}, we can access a richer data set. An enriched data pool can facilitate more precise classification and recognition for various perceptual tasks, including identifying the point of contact of the robot with obstacles or classifying different types of terrains and obstructions.

The resulting digital-twinned perception can serve as a substantial basis for learning high-level soft robot control, such as reinforcement learning. Training models in simulations across diverse configurations can enable the development of a versatile pipeline that mitigates extensive experimentation in the real world. Our framework can facilitate computationally efficient sim-to-real transfer of the learned control strategy.

\bibliography{reference.bib}
\bibliographystyle{ieeetr}

\begin{IEEEbiography}[{\includegraphics[width=1in,height=1.25in,clip,keepaspectratio]{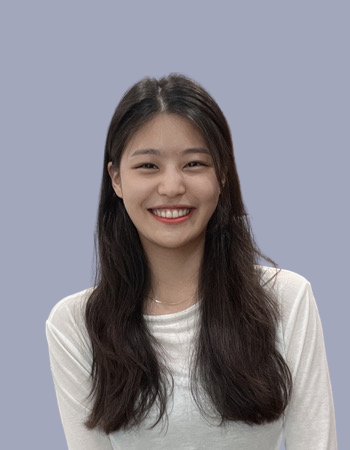}}]{Charee Park}
received her B.S. and M.S. degrees in mechanical engineering from Korea Advanced Institute of Science and Technology (KAIST), Daejeon, Republic of Korea, in 2021 and 2023, respectively. Her research interests include soft robotics, manipulation, and robot learning. 
\end{IEEEbiography}

\begin{IEEEbiography}[{\includegraphics[width=1in,height=1.25in,clip,keepaspectratio]{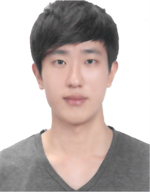}}]{Hyunkyu Park}
received his B.S. and M.S. degrees in mechanical engineering from Korea Advanced Institute of Science and Technology (KAIST), Daejeon, Republic of Korea, in 2017 and 2019, respectively. He has been working toward his Ph.D. degree at KAIST. His research interests include soft robotics, deep learning, and wearable robotics. 
\end{IEEEbiography}

\begin{IEEEbiography}[{\includegraphics[width=1in,height=1.25in,clip,keepaspectratio]{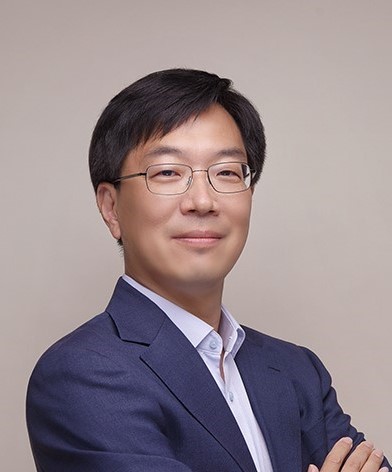}}]{Jung Kim}
received his B.S. and M.S. degrees in mechanical engineering of Korea Advanced Institute of Science and Technologies (KAIST), Korea, in 1991 and 1993, respectively. He also holds a Ph. D. degree in mechanical engineering from Massachusetts Institute of Technology (MIT), Cambridge, US, which he earned in 2003. He has joined the department of mechanical engineering at KAIST in 2004 and is currently a professor in the same department. His current research interests include medical robotics, human-robot interaction (HRI), bio-mechanical systems, and assistive robotics.
\end{IEEEbiography}

\newpage
\vfill

\end{document}